\documentclass{article}

\usepackage{times}
\usepackage{graphicx} 
\usepackage{subfigure} 

\usepackage{natbib}

\usepackage{algorithm}
\usepackage{algorithmic}

\usepackage{hyperref}

\usepackage[accepted]{icml2017} 

\usepackage[T1]{fontenc}
\usepackage[font=small,labelfont=bf,tableposition=top]{caption}
\DeclareCaptionLabelFormat{andtable}{#1~#2  \&  \tablename~\thetable}

\icmltitlerunning{Forward and Reverse Gradient-Based Hyperparameter Optimization}

\usepackage{amsmath}
\usepackage{amssymb}

\usepackage{clrscode3e}
\usepackage{varwidth}
\def\RSet{\mathbb{R}}

\usepackage[status=draft,margin=false,inline=true,index=true]{fixme}

\newcommand{\Lagr}{{\cal L}}

\newcommand{\vx}{x}
\newcommand{\mw}{W}
\newcommand{\vlambda}{{\lambda}}

\newcommand{\mc}{C}

\newcommand{\vb}{b}
\newcommand{\luca}[1]{\textcolor{green}{#1}}

\def\transpose#1{#1^\intercal}

\usepackage[utf8]{inputenc}

\begin{document} 

\twocolumn[
\icmltitle{Forward and Reverse Gradient-Based Hyperparameter Optimization}



\icmlsetsymbol{equal}{*}

\begin{icmlauthorlist}
\icmlauthor{Luca Franceschi}{iit,ucl}
\icmlauthor{Michele Donini}{iit}
\icmlauthor{Paolo Frasconi}{fi}
\icmlauthor{Massimiliano Pontil}{iit,ucl}
\end{icmlauthorlist}

\icmlaffiliation{iit}{Computational Statistics and Machine Learning, Istituto Italiano di Tecnologia, Genoa, Italy}
\icmlaffiliation{ucl}{Department of Computer Science, University College London, UK}
\icmlaffiliation{fi}{Department of Information Engineering, Università degli Studi di Firenze, Italy}

\icmlcorrespondingauthor{Luca Franceschi}{luca.franceschi@iit.it}

\icmlkeywords{hyperparameter optimization}

\vskip 0.3in
]



\printAffiliationsAndNotice{}  

\begin{abstract} 
  We study two procedures (reverse-mode and forward-mode) for computing the
  gradient of the validation error with respect to the hyperparameters
  of any iterative learning algorithm such as stochastic
  gradient descent. These procedures mirror two methods of computing gradients
  for recurrent neural networks and have different trade-offs in terms
  of running time and space requirements. 
  Our formulation of the reverse-mode procedure is linked to previous work 
  by \citet{maclaurin2015gradient} but does not require reversible dynamics.
  The forward-mode procedure is suitable for real-time
  hyperparameter updates, which may significantly speed up 
  hyperparameter optimization on large datasets.
  We present  experiments
   on data cleaning and  on learning task interactions.
  We also present one large-scale experiment where the use of 
  previous gradient-based methods would be prohibitive.
\end{abstract}


\section{Introduction}

The increasing complexity of machine learning algorithms has driven a large amount of research in the area of hyperparameter
optimization (HO) --- see, e.g.,~\cite{hutter_beyond_2015} for a
review. The core idea is relatively simple: given a measure of
interest (e.g. the misclassification error) HO methods use a
validation set to construct a \textit{response function} of the
hyperparameters (such as the average loss on the validation set)
and explore the hyperparameter space to seek an optimum. 

Early approaches based on grid search quickly become impractical as the 
number of hyperparameters grows and are even outperformed by random search~\cite{bergstra2012random}. Given the high computational 
cost of evaluating the response function, Bayesian optimization approaches 
provide a natural framework and have been extensively studied in this context~
\cite{snoek_practical_2012,swersky_multi-task_2013,snoek_scalable_2015}.
Related and faster sequential model-based optimization methods have been proposed
using random forests~\cite{hutter_sequential_2011} and tree Parzen
estimators~\cite{bergstra_algorithms_2011}, scaling up to a few
hundreds of hyperparameters~\cite{bergstra_making_2013}. 

In this paper, we follow an alternative direction, where 
gradient-based algorithms are used to optimize the performance on a validation set with respect to the hyperparameters~\cite{bengio2000gradient,larsen1996design}. In this setting, the validation
error should be evaluated at a minimizer of the training objective. However, in many current learning systems such as deep learning, the minimizer is only approximate. \citet{domke2012generic} specifically considered running
an iterative algorithm, like gradient descent or momentum, for a given
number of steps, and subsequently computing the gradient of the
validation error by a back-optimization
algorithm. \citet{maclaurin2015gradient} considered reverse-mode
differentiation of the response function. They suggested the idea of
reversing parameter updates to achieve space efficiency, proposing an
approximation capable of addressing the associated loss of information
due to finite precision
arithmetics. \citet{pedregosa_hyperparameter_2016} proposed the use of
inexact gradients, allowing hyperparameters to be updated before
reaching the minimizer of the training objective. Both
\cite{maclaurin2015gradient} and \cite{pedregosa_hyperparameter_2016}
managed to optimize a number of hyperparameters in the order of one thousand. 

In this paper, we illustrate two alternative approaches to compute the hypergradient 
(i.e., the gradient of the response function), which 
have different trade-offs in terms of running time and space requirements.
One approach is based on a Lagrangian formulation associated with
the parameter optimization dynamics. 
It encompasses the reverse-mode differentiation (RMD) approach used by \citet{maclaurin2015gradient}, where the dynamics corresponds to stochastic gradient descent with momentum. We do not assume reversible parameter optimization dynamics.  A well-known drawback of RMD is its space complexity: we need 
to store the whole trajectory of training iterates
in order to compute the hypergradient.  An alternative approach that we consider overcomes this
problem by computing the hypergradient in forward-mode and it is
efficient when the number of hyperparameters is much smaller than the 
number of parameters. To the best of our knowledge, the forward-mode
has not been studied before in this context.

As we shall see, these two approaches have a direct correspondence 
to two classic alternative ways of computing gradients for recurrent
neural networks (RNN) \cite{pearlmutter_gradient_1995}: 
the Lagrangian (reverse) way corresponds to
back-propagation through time~\cite{werbos90}, while
the forward way corresponds to real-time
recurrent learning (RTRL) \cite{williams_learning_1989}. As RTRL
allows one to update parameters after each time step, the forward
approach is suitable for real-time hyperparameter updates, which may
significantly speed up the overall hyperparameter optimization
procedure in the presence of large datasets. We give experimental
evidence that the real-time approach is efficient enough 
to allow for the automatic tuning of crucial hyperparameters in a deep learning model.  
In our experiments, we also explore constrained hyperparameter optimization, showing that it can be used effectively to detect noisy examples and to discover the relationships between different learning tasks. 



The paper is organized in the following manner. In Section~\ref{Sec:HO} we introduce the problem under study. In Section~\ref{sec:reverse} we derive the
reverse-mode computation. 
In Section~\ref{sec:forward} we present the forward-mode computation of the hypergradient, and in Section~\ref{sec:real-time} we introduce the idea of real-time hyperparameter updates.
In Section \ref{sec:complexity} 
we discuss the time and space complexity of these methods. In Section
\ref{sec:exps} we present empirical results with both
algorithms. Finally in Section \ref{sec:discussion} we discuss our
findings and highlight directions of future research.



\section{Hyperparameter Optimization}
\label{Sec:HO}
We focus on training procedures based on the optimization of an objective
function $J(w)$ with respect to $w$ (e.g. the regularized average training
loss for a neural network with weights $w$).
We see the training procedure by stochastic gradient descent (or one of its variants like momentum,
RMSProp, Adam, etc.) as a dynamical system with a state $s_t\in\RSet^d$ that collects 
weights and possibly accessory variables such as velocities and accumulated 
squared gradients. The dynamics are defined by the system of equations
\begin{equation}
  \label{eq:dynamics}
  s_t = \Phi_t(s_{t-1},\lambda) \ \ t=1,\dots,T
\end{equation}
where $T$ is the number of iterations, $s_0$ contains initial weights
and initial accessory variables, and, for every $t \in \{1,\dots,T\}$,
$$
\Phi_t : (\RSet^d \times \RSet^m )\rightarrow \RSet^d
$$
is a smooth mapping that represents the operation performed by the $t$-th step of 
the optimization algorithm (i.e. on mini-batch $t$). Finally, $\lambda\in\RSet^m$ is 
the vector of hyperparameters that we wish to tune.

As simple example of these dynamics occurs when 
training a neural network by gradient descent with momentum (GDM), in which case $s_t=(v_t,w_t)$ and
\begin{equation}
  \label{eq:momentum}
     \begin{array}{lcl}
       v_t & = &\mu v_{t-1} + \nabla J_t(w_{t-1}) \\
      w_t & = & w_{t-1} - \eta(\mu v_{t-1} - \nabla J_t(w_{t-1}))
    \end{array} 
    \end{equation}
where $J_t$ is the objective associated with the $t$-th mini-batch, $\mu$ is the rate and $\eta$ is the momentum. In
this example, $\lambda=(\mu,\eta)$.


Note that the iterates $s_1,\dots,s_T$ implicitly depend on the vector of hyperparameters $\lambda$. Our goal is to optimize the hyperparameters according to 
a certain error function $E$ evaluated at the last iterate $s_T$.
Specifically, we wish to solve the problem
\begin{equation}
  \min_{\lambda\in \Lambda} f(\lambda)
  \label{eq:minval}
\end{equation}
where the set $\Lambda\subset \mathbb{R}^m$ 
incorporates constraints on the hyperparameters, and the response function $f : \RSet^m \rightarrow \RSet$ is defined at $\lambda \in \RSet^m$ as
\begin{equation}
  f(\lambda) = E(s_T(\lambda)).
  \label{resp-func}
\end{equation}
We highlight the generality of the framework. The vector of hyperparameters $\lambda$ may 
include components associated with the training objective, and components associated with the 
iterative algorithm. For example, the training objective may depend on hyperparameters used to design the loss function as well as multiple regularization parameters. Yet other components of $\lambda$ may be associated with the space of functions used to fit the training objective (e.g. number of layers and weights of a neural network, parameters associated with the kernel function used within a kernel based method, etc.). The validation error $E$ can in turn be of different kinds. The simplest example is to choose $E$ as the average of a loss function over a validation set. 
We may however consider multiple validation objectives, in that the hyperparameters associated with the iterative algorithm ($\mu$ and $\gamma$ in the case of momentum mentioned above) may be optimized using the training set, whereas the regularization parameters would typically require a validation set, which is distinct from the training set (in order to avoid over-fitting).


\section{Hypergradient Computation}
In this section, we review the reverse-mode computation of
the gradient of the response function (or hypergradient)
under a Lagrangian perspective and introduce a forward-mode strategy. These procedures correspond to the reverse-mode and the forward-mode algorithmic 
differentiation schemes~\cite{griewank_evaluating_2008}.
We finally introduce a real-time version of the forward-mode procedure.

\subsection{Reverse-Mode}
\label{sec:reverse}
The reverse-mode computation leads to an algorithm closely related to
the one presented in \cite{maclaurin2015gradient}.
A major difference with respect to their work is that
we do not require the mappings $\Phi_t$ defined in
Eq.~(\ref{eq:dynamics}) to be invertible.
We also note that the reverse-mode calculation is structurally identical to back-propagation through time~\cite{werbos90}.  

\begin{algorithm}[t]
  \caption{\proc{Reverse-HG}}
  \label{alg:ho-reverse}
  \begin{algorithmic}
    \STATE {\bfseries Input:} $\lambda$ current values of the
    hyperparameters, $s_0$ initial optimization state
    \STATE {\bfseries Output:} Gradient of validation error w.r.t. $\lambda$
    \FOR{$t=1$ {\bfseries to} $T$}
    \STATE $s_t\gets \Phi_t(s_{t-1},\lambda)$
    \ENDFOR
    \STATE $\alpha_T \gets \nabla E(s_T)$
    \STATE $g\gets 0$
    \FOR {$t=T-1$ {\bfseries downto} $1$}
    \STATE       $g \gets g + \alpha_{t+1} B_{t +1} $ 
    \STATE $\alpha_t \gets  \alpha_{t+1}A_{t+1}$
    \ENDFOR
    \STATE {\bf return} $g$
  \end{algorithmic}
\end{algorithm}

We start by reformulating problem \eqref{eq:minval} as the constrained optimization problem
\begin{equation}
  \begin{array}{cl}
    \min\limits_{\lambda,s_1,\dots,s_T} & E(s_T) \\ 
   \text{~~~~~subject to} & s_t =  \Phi_t(s_{t-1},\lambda),~t \in \{1,\dots,T\}.
  \end{array}
  \label{eq:ddd}
\end{equation}
This formulation closely follows a classical Lagrangian approach used
to derive the back-propagation algorithm~\cite{lecun_theoretical_1988}. 
Furthermore, the framework naturally allows one to incorporate constraints on the hyperparameters.

The Lagrangian of problem \eqref{eq:ddd} is
\begin{equation}
  {\cal L}(s,\lambda,\alpha) = E(s_T) + \sum_{t=1}^T \alpha_t( \Phi_t(s_{t-1},\lambda)- s_t)
\end{equation}
where, for each $t \in \{1,\dots,T\}$, $\alpha_t  \in {\mathbb R}^d$ is a row vector of Lagrange multipliers associated with the $t$-th step of the dynamics.

The partial derivatives of the Lagrangian are given by
\begin{eqnarray}
\frac{\partial {\Lagr} }{\partial \alpha_t} & \hspace{-.13truecm}=  \hspace{-.13truecm}& \Phi_t(s_{t-1},\lambda)- s_t,~~~t \in \{1,\dots,T\}
\label{eq:L1} \\
\frac{\partial {\Lagr} }{\partial s_t} 
&  \hspace{-.13truecm}=  \hspace{-.13truecm}& \alpha_{t+1} A_{t+1} - \alpha_t,~~~~\hspace{.05truecm}t \in 
\{1,\dots,T \hspace{-.05truecm}- \hspace{-.05truecm}1\} \label{eq:L2} \\
\frac{\partial {\Lagr} }{\partial s_T} & \hspace{-.13truecm} =  \hspace{-.13truecm}& \nabla E(s_T) - \alpha_T \label{eq:L2-T} \\
\frac{\partial {\Lagr} }{\partial \lambda} &  \hspace{-.13truecm}=  \hspace{-.13truecm}& \sum_{t=1}^T 
\alpha_t B_t,
\label{eq:L3}
\end{eqnarray}
where for every $ t \in \{1,\dots,T\}$, we define the matrices
\begin{equation}
A_t  = \frac{\partial \Phi_t(s_{t-1},\lambda)}{\partial s_{t-1}}, \,\,\,\,
B_t  = \frac{\partial \Phi_t(s_{t-1},\lambda)}{\partial \lambda}.
\label{eq:AtBt}
\end{equation}
Note that $A_t\in \mathbb{R}^{d \times d}$ and $B_t\in\mathbb{R}^{d \times m}$.

The optimality conditions are then obtained by setting each derivative to zero. In particular, setting the right hand side of Equations~\eqref{eq:L2} and \eqref{eq:L2-T} to zero gives
\begin{equation}
\alpha_t =
\begin{cases}
 \nabla E(s_T)&\text{if } t = T,\\ \nonumber\\
  \nabla E(s_{T}) A_{T} \cdots A_{t+1} & \text{if } t \in\{1,\dots,T \hspace{-.05truecm}- \hspace{-.05truecm}1\} .
\end{cases}
\label{eq:LLL}
\end{equation}
Combining these equations with Eq.~\eqref{eq:L3} we obtain that
$$
\frac{\partial {\Lagr} }{\partial \lambda} =  \nabla E(s_T) \sum_{t=1}^T \left(  \prod_{s=t+1}^T A_{s} \right)  B_t.
$$
As we shall see this coincides with the expression for the gradient of $f$ in Eq.~\eqref{eq:grad} derived in the next section. Pseudo-code of \proc{Reverse-HG} is presented in Algorithm \ref{alg:ho-reverse}.

\begin{algorithm}[tb]
  \caption{\proc{Forward-HG}}
  \label{alg:ho-forward}
  \begin{algorithmic}
    \STATE {\bfseries Input:} $\lambda$ current values of the
    hyperparameters, $s_0$ initial optimization state
    \STATE {\bfseries Output:} Gradient of validation error w.r.t. $\lambda$
    \STATE $Z_0 \gets 0$
    \FOR{$t=1$ {\bfseries to} $T$}
    \STATE $s_t\gets \Phi_t(s_{t-1},\lambda)$
    \STATE $Z_t \gets A_t Z_{t-1} + B_t$
    \ENDFOR
    \STATE {\bf return} $\nabla E(s) Z_T $
  \end{algorithmic}
\end{algorithm}

\subsection{Forward-Mode}
\label{sec:forward}
The second approach to compute the hypergradient appeals to the chain rule for the derivative of composite functions, to obtain that the gradient of $f$ at $\lambda$ satisfies\footnote{Remember that the gradient of a scalar function is a row vector.}
\begin{equation}
\nabla f(\lambda) =  \nabla E(s_T) \frac{d s_T}{d
  \lambda}
\label{eq:1}
\end{equation}
where 
$\frac{d s_T}{d \lambda}$ is the $d \times m$ matrix formed by the
total derivative of the components of $s_T$ (regarded as rows) with
respect to the components of $\lambda$ (regarded as columns).

Recall that $s_t = \Phi_t(s_{t-1},\lambda)$. The operators $\Phi_t$
depend on the hyperparameter $\lambda$ both directly by its
expression and indirectly through the state $s_{t-1}$. Using again the
chain rule we have, for every $t \in \{1,\dots,T\}$, that
\begin{equation}
  \frac{d s_t}{d \lambda} =
\frac{ \partial \Phi_t(s_{t-1},\lambda)}{\partial s_{t-1}} \frac{d s_{t-1}}{d \lambda} 
+ \frac{\partial \Phi_t(s_{t-1},\lambda)}{\partial \lambda}\,. 
\label{eq:rec0}
\end{equation}
Defining $Z_t =  \frac{d s_t}{d \lambda}$ for every $ t \in \{1,\dots,T\}$ and recalling Eq.~\eqref{eq:AtBt}, we can rewrite Eq.~\eqref{eq:rec0} as the recursion
\begin{equation}
  Z_t =  A_t  Z_{t-1} + B_t,~~~t \in \{1,\dots,T\}.
\label{eq:rec1}
\end{equation}
Using Eq.~\eqref{eq:rec1}, we obtain that
\begin{eqnarray}
\nonumber
\nabla f(\lambda) & \hspace{-.75truecm} =   \hspace{-.75truecm}&   \nabla E(s_T) Z_T \\ \nonumber
~ & \hspace{-.75truecm} =   \hspace{-.75truecm} & \nabla E(s_T) ( A_T Z_{T-1} + B_T) \\ \nonumber
~ & \hspace{-.75truecm} =  \hspace{-.75truecm} & \nabla E(s_T) (A_T A_{T-1} Z_{T-2} + A_T B_{T-1}  + B_T) \\ \nonumber
 &\vdots & \\
~& \hspace{-.75truecm} = \hspace{-.75truecm} &  
\nabla E(s_T) \sum_{t=1}^T \left(  \prod_{s=t+1}^T A_{s} \right)  B_t.
\label{eq:grad}
\end{eqnarray}
Note that the recurrence~(\ref{eq:rec1}) on the Jacobian matrix 
is structurally identical to the recurrence in the RTRL
procedure described in \citep[eq.~(2.10)]{williams_learning_1989}.


From the above derivation it is apparent that $\nabla f(\lambda)$ can
be computed by an iterative algorithm which runs in parallel to the
training algorithm. Pseudo-code of \proc{Forward-HG} is presented in Algorithm \ref{alg:ho-forward}. At first sight, the computation of the terms in the right hand side of Eq.~\eqref{eq:rec1} seems prohibitive. However, in Section \ref{sec:complexity} we observe that if
$m$ is much smaller than $d$, the computation can be done efficiently. 



\subsection{Real-Time Forward-Mode}
\label{sec:real-time}
For every $t \in \{1,\dots,T\}$  let $f_t: \RSet^m \rightarrow \RSet$ be the response function
at time $t$: $f_t(\lambda) = E(s_t(\lambda))$. Note that $f_T$ coincides with
the definition of the response function in Eq.~\eqref{resp-func}.
A major difference between \proc{Reverse-HG} and \proc{Forward-HG} is that the
\textit{partial} hypergradients
\begin{equation}
  \nabla f_t(\lambda) = \frac{dE(s_t)}{d\lambda} =  \nabla E(s_t) Z_t 
  \label{eq:partial-hypergradient}
\end{equation}
are available in the second procedure at each time step $t$ and not only at the end. 

The availability of partial hypergradients is significant since we are allowed to update
hyperparameters several times in a single optimization epoch, without
having to wait until time $T$. 
This is reminiscent of the real-time
updates suggested by Williams \& Zipser (\citeyear{williams_learning_1989})
for RTRL. 
The real-time approach may be suitable in the case of a data stream
(i.e. $T=\infty$), where \proc{Reverse-HG} would be hardly
applicable. 
Even in the case of finite  (but large)
datasets it is possible to perform one
hyperparameter update after a hyper-batch
of data (i.e. a set of minibatches) has been processed.
Algorithm~\ref{alg:ho-forward} can be easily modified to yield a partial
hypergradient when $t \mod \Delta=0$ (for some hyper-batch size $\Delta$) and
letting $t$ run from $1$ to $\infty$, reusing examples in a circular or random way.
We use this strategy in the phone recognition experiment reported in
Section~\ref{sec:speech}.





\section{Complexity Analysis}
\label{sec:complexity}
We discuss the time and space complexity of Algorithms~\ref{alg:ho-reverse}
and~\ref{alg:ho-forward}. 
We begin by recalling some basic results from the
algorithmic differentiation (AD) literature.

Let $F:\RSet^n\mapsto\RSet^p$ be a differentiable function and suppose
it can be evaluated in time $c(n,p)$ and requires space $s(n,p)$. Denote by
$J_F$ the $p\times n$ Jacobian matrix of $F$. Then the following facts
hold true~\cite{griewank_evaluating_2008} (see 
also~\citet{baydin_automatic_2015} for a shorter account):
\begin{enumerate}
\item[(i)] For any vector $r\in\RSet^n$, the product $J_Fr$ can be
  evaluated in time $O(c(n,p))$ and requires space $O(s(n,p))$ using
  forward-mode AD.\label{item:AD-jac-vec}
\item[(ii)]  For any vector $q\in\RSet^p$, the product $\transpose{J}_F q$ has
  time and space complexities $O(c(n,p))$ using reverse-mode
  AD.\label{item:AD-jact-vec}
  
\item[(iii)]  As a corollary of item (i), the whole $J_F$
  can be computed in time $O(n c(n,p))$ and requires space $O(s(n,p))$ using
  forward-mode AD (just use unitary vectors $r=e_i$ for
  $i=1,\dots,n$).\label{item:AD-jac-forward}
\item[(iv)]  Similarly, $J_F$ can be computed in time $O(p c(n,p))$ and 
  requires space $O(c(n,p))$ using reverse-mode AD.\label{item:AD-jac-reverse}
\end{enumerate}

Let $g(d,m)$ and $h(d,m)$ denote time and
space, respectively, required to evaluate the update map $\Phi_t$
defined by Eq.~(\ref{eq:dynamics}). Then the response function
$f:\RSet^m\mapsto\RSet$ defined in Eq.~(\ref{eq:minval}) can be
evaluated in time $O(Tg(d,m))$ (assuming the time required to compute
the validation error $E(\lambda)$ does not affect the
bound\footnote{This is indeed realistic since the number of validation
  examples is typically lower than the number of training
  iterations.})  and requires space $O(h(d,m))$ since variables $s_t$ may be
overwritten at each iteration.
Then, a direct application of Fact (i) above shows that
Algorithm~\ref{alg:ho-forward} runs in time $O(T m g(d,m))$ and space
$O(h(d,m))$.  The same results can also be obtained by noting that
in Algorithm~\ref{alg:ho-forward} the product $A_tZ_{t-1}$ requires
$m$ Jacobian-vector products, each costing $O(g(d,m))$ (from Fact (i)), while computing the Jacobian $B_t$ takes time $O(mg(d,m))$ (from Fact (iii)).

Similarly, a direct application of Fact (ii) shows that
Algorithm~\ref{alg:ho-reverse} has both time and space complexities
$O(T g(d,m))$. Again the same results can be obtained by noting that
$\alpha_{t+1}A_{t_1}$ and $\alpha_t B_t$ are
transposed-Jacobian-vector products that in reverse-mode take both time $O(g(d,m))$ (from Fact (ii)).
Unfortunately in this case variables
$s_t$ cannot be overwritten, explaining the much higher space
requirement.

As an example, consider training a neural network with $k$
weights\footnote{This includes linear SVM and logistic regression as
  special cases.}, using classic iterative optimization algorithms
such as SGD (possibly with momentum) or Adam, where the
hyperparameters are just learning rate and momentum terms. In this
case, $d=O(k)$ and $m=O(1)$. Moreover, $g(d,m)$ and $h(d,m)$ are both
$O(k)$. As a result, Algorithm~\ref{alg:ho-reverse} runs in time and
space $O(Tk)$, while Algorithm~\ref{alg:ho-forward} runs in time
$O(Tk)$ and space $O(k)$, which would typically make a dramatic
difference in terms of memory requirements.


\section{Experiments}
\label{sec:exps}
In this section, we present numerical simulations with the proposed methods.
All algorithms were implemented in
TensorFlow and the software package used to reproduce our
experiments is available\footnote{A newer version of the package is available at \url{https://github.com/lucfra/FAR-HO}.} at \url{https://github.com/lucfra/RFHO}.
In all the experiments, hypergradients were used inside
the Adam algorithm~\cite{kingma_adam:_2014} in order to 
minimize the response function.





\subsection{Data Hyper-cleaning}
\label{sec:hypercleaning}
The goal of this experiment is to highlight one potential advantage of
constraints on the hyperparameters. Suppose we have a dataset with
label noise and due to time or resource constraints we can only afford
to cleanup (by checking and correcting the labels)
a subset of the available data. Then we may use the cleaned
data as the validation set, the rest as the training set, and
assign one hyperparameter to each training
example.  By putting a sparsity constraint on the vector
of hyperparameters $\lambda$, we
hope to bring to zero the influence of noisy examples, in order to
generate a better model. While this is the same kind of data sparsity
observed in support vector machines (SVM),
our setting aims to get rid of erroneously labeled
examples, in contrast to SVM which puts zero weight on redundant
examples. 
Although this experimental setup
does not necessarily reflect a
realistic scenario, it aims to test the ability of our HO method to
effectively make use of constraints on the hyperparameters\footnote{We note that a related approach based  on reinforcement learning is presented in \cite{fan2017learning}.}

We instantiated the above setting with a balanced subset of $N=20000$
examples from the MNIST dataset, split into three subsets:
$\mathcal{D}_{\operatorname{tr}}$ of $N_{\operatorname{tr}} = 5000$ training examples, $\mathcal{V}$
of $N_{\operatorname{val}} = 5000$ validation examples and 
a test set containing the
remaining samples. 
Finally, we corrupted the labels of $2500$
training examples, selecting a random subset
$\mathcal{D}_{f} \subset \mathcal{D}_{\operatorname{tr}}$.

We considered a plain softmax regression model with parameters $\mw$
(weights) and $b$ (bias). The error of a model $(\mw,\vb)$ on an example
$\vx$ was evaluated by using the cross-entropy $\ell(\mw,\vb,\vx)$ both
in the training objective function, $E_{\mathrm{tr}}$, and in the validation
one, $E_{\operatorname{val}}$. 
We added in $E_{\operatorname{tr}}$ an hyperparameter vector
$\vlambda \in [0,1]^{N_{\operatorname{tr}}}$ that weights each example 
in the training phase, i.e.
$E_{\operatorname{tr}}(\mw,\vb) = \frac{1}{N_{\operatorname{tr}}} \sum_{i \in \mathcal{D}_{\operatorname{tr}}}
\lambda_i \ell(\mw,\vb,\vx_i)$.

According to the general HO framework, we fit the parameters
$(\mw,\vb)$ to minimize the training loss and the hyperparameters
$\vlambda$ to minimize the validation error.  The sparsity constraint
was implemented by bounding the $L1$-norm of $\lambda$, resulting
in the optimization problem
\[
\min_{\vlambda\in \Lambda}~ E_{\operatorname{val}}(\mw_{T},\vb_{T})~~~~~~~~(P_{HO}) 
 \]
where $\Lambda = \{\lambda:\lambda\in [0,1]^{N_{\operatorname{tr}}},  \| \vlambda \|_1 \leq R\}
$ and $(\mw_T,b_T)$ are the parameters obtained after $T$ iterations of gradient descent on the training objective.
Given the high dimensionality of $\lambda$,
we solved $(P_{HO})$ iteratively computing the hypergradients with \proc{Reverse-HG} method and 
projecting Adam updates on the set $\Lambda$.

We are interested in comparing the following three test set accuracies:
\begin{itemize}
\item Oracle: the accuracy of the minimizer of $E_{{\operatorname{tr}}}$
  trained on clean examples only, i.e.
  $(\mathcal{D}_{{\operatorname{tr}}} \setminus \mathcal{D}_{f}) \cup \mathcal{V}$;
  this setting is effectively taking advantage of an oracle that tells
  which examples have a wrong label;
\item Baseline: the accuracy of the minimizer of $E_{\operatorname{tr}}$
  trained on all available data $\mathcal{D} \cup \mathcal{V}$;
\item DH-R: 
the accuracy of the data hyper-cleaner with a
  given value of the $L1$ radius, $R$. In this case, we first optimized
  hyperparameters and then constructed a cleaned training set
  $\mathcal{D}_c\subset\mathcal{D}_{\mathrm{tr}}$ (keeping examples with
  $\lambda_i>0$); we finally trained on
  $\mathcal{D}_c \cup \mathcal{V}$.
\end{itemize}

\begin{table}[h]
  \caption{ \label{tab:mnist_flipped} Test accuracies for the
    baseline, the oracle, and DH-R 
    for four different values of $R$. The reported $F_1$ measure is the
    performance of the hyper-cleaner in correctly identifying the
    corrupted training examples.}
  \centering
  \begin{tabular}{lcc}
    \hline
    \abovespace\belowspace
    & Accuracy \% & $F_1$  \\ \hline
    Oracle            & 90.46   & 1.0000     \\
    Baseline            & 87.74   & -      \\
    ${\textrm{DH-}1000}$ & 90.07   & 0.9137 \\
    ${\textrm{DH-}1500}$ & 90.06   & 0.9244 \\
    ${\textrm{DH-}2000}$ & 90.00      & 0.9211 \\
    ${\textrm{DH-}2500}$ & 90.09   & 0.9217 \\  \hline 
%
  \end{tabular}
\end{table}

We are also interested in evaluating the ability of the hyper-cleaner to detect noisy samples.
Results are shown in Table~\ref{tab:mnist_flipped}. The data
hyper-cleaner is robust with respect to the choice of $R$ and is able
to identify corrupted examples, recovering a model that has almost the
same accuracy as a model produced with the help of an oracle.

\begin{figure}[h]
  \includegraphics[width=8cm]{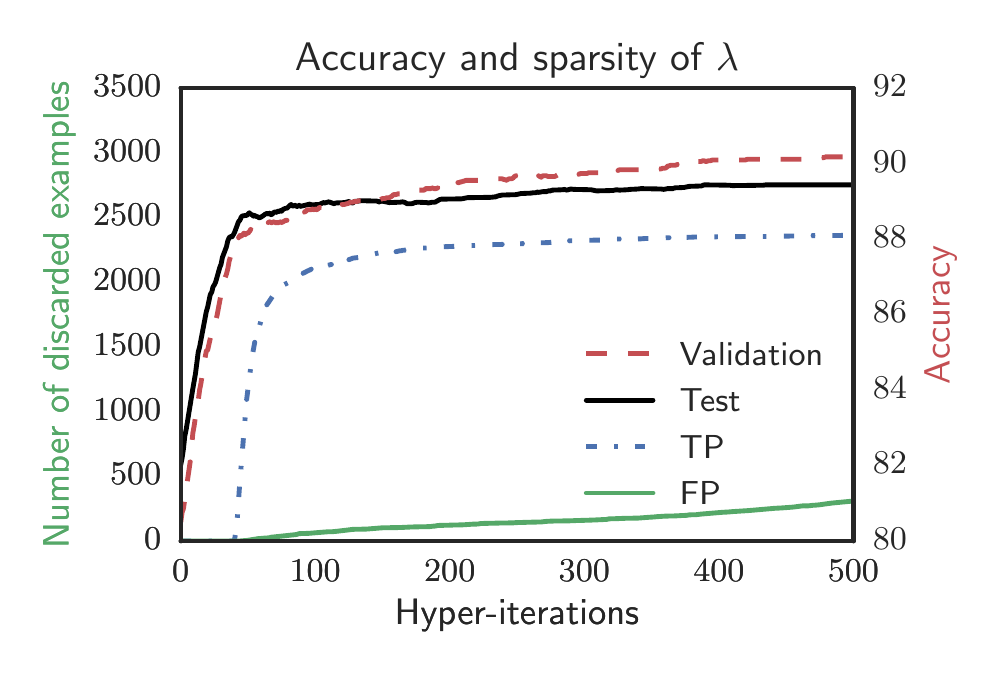}
  \caption{Right vertical axis: accuracies of DH-1000 on validation and test sets.  
 Left vertical axis: number of discarded examples among noisy (True Positive, TP) and clean (False Positive, FP) ones.}   \label{fig:mnist_flipped} 
\end{figure}

Figure~\ref{fig:mnist_flipped} shows how the accuracy of
${\textrm{DH-}1000}$ improves with the number of
hyper-iterations and the progression of the amount of discarded examples.
The data hyper-cleaner starts by discarding mainly corrupted examples, and while the optimization proceeds,  it begins to remove also a portion of cleaned one.
Interestingly, the
test set accuracy continues to improve even when some of the clean examples are discarded.

\subsection{Learning Task Interactions}
\label{sec:vision}
This second set of experiments is in the multitask learning (MTL) context, 
where the goal is to find simultaneously the model of 
multiple related tasks.
Many MTL methods require that a task interaction matrix 
is given as input to the
learning algorithm. However, in real applications, this matrix is often unknown 
and it is interesting to learn it from data. Below, we show that our framework can be naturally applied to learning the task relatedness matrix.

We used CIFAR-10 and CIFAR-100~\cite{krizhevsky2009learning}, two object recognition datasets with $10$ and $100$ classes, respectively. As features we employed the pre-activation of the second last layer of Inception-V3 model trained on ImageNet\footnote{Available at \url{tinyurl.com/h2x8wws}
}. From CIFAR-10, we extracted $50$ examples as training set,
different $50$ examples as validation set and the remaining for testing. 
From CIFAR-100, we selected $300$ examples as training set, $300$ as validation set and the remaining for testing. Finally, we used a one-hot encoder of the labels obtaining a set of labels in $\{0,1\}^{K}$ ($K=10$ or $K=100$).

The choice of small training set sizes is due to the strong discriminative power of the selected features. In fact, using larger sample sizes would not allow to appreciate the advantage of MTL. 
In order to leverage information among the different classes, we employed a 
multitask learning (MTL) regularizer  \citep{evgeniou2005learning} 
$$
\Omega_{C,\rho}(W)= \sum_{j,k = 1}^{K} C_{j,k}\| w_j - w_k \|_2^2 + \rho \sum_{k=1}^K \|w_k\|^2,
$$ 
where $w_k$ are the weights for class $k$, $K$ is the number of classes, and
the symmetric non-negative matrix $C$ models the interactions between the classes/tasks. 
We used a regularized training error defined as $E_{\operatorname{tr}}(W) = \sum_{i \in \mathcal{D}_{\operatorname{tr}} } \ell(W x_i + b,y_i) + \Omega_{C,\rho}(W)$ where $\ell(\cdot,\cdot)$ is the categorical cross-entropy and $b=(b_1,\dots,b_K)$ is the vector of thresholds associated with each linear model. We wish solve the following optimization problem: 
$$
\min\big\{ E_{\operatorname{val}}(W_T,b_T) ~\text{ subject to }~  \rho \geq 0,~C=\transpose{C},~C \geq 0\big\},
$$
where $(\mw_T,b_T)$ is the $T$-th iteration obtained by running gradient 
descent with momentum (GDM) on the training objective. 
We solve this problem using \proc{Reverse-HG} and optimizing the 
hyperparameters by projecting Adam updates on the set
$\{(\rho, C) : \rho \geq 0,~C=\transpose{C},~C \geq 0 \}$. 
We compare the following methods:
\begin{itemize}
\item SLT: single task learning, i.e. $C = 0$, using a validation set to tune the optimal value of $\rho$ for each task;
\item NMTL: we considered the naive MTL scenario in which the tasks are equally related, that is $C_{j,k}= a$ for every $1\leq j,k \leq K$. In this case we learn the two non-negative hyperparameters $a$ and $\rho$;
\item HMTL: our hyperparameter optimization method \proc{Reverse-HG} to tune $C$ and $\rho$;
\item HMTL-S: Learning the matrix $\mc$ with only few examples per class could bring the discovery of spurious relationships. 
We try to remove this effect by imposing the constraint that $\sum_{j,k} C_{j,k} \leq R$, where\footnote{We observed that $R=10^{-4}$ yielded very similar results.} $R = 10^{-3}$. In this case, Adam updates are projected onto the set $\{(\rho, \mc) : \rho \geq 0,~C=\transpose{C},~ C \geq 0, ~\sum_{j,k} C_{j,k} \leq R\}$.
\end{itemize}



  \begin{figure}[!t]
  	\centering
    \includegraphics[width=3.98cm, trim={0cm 0cm 0.0cm 0cm},clip]{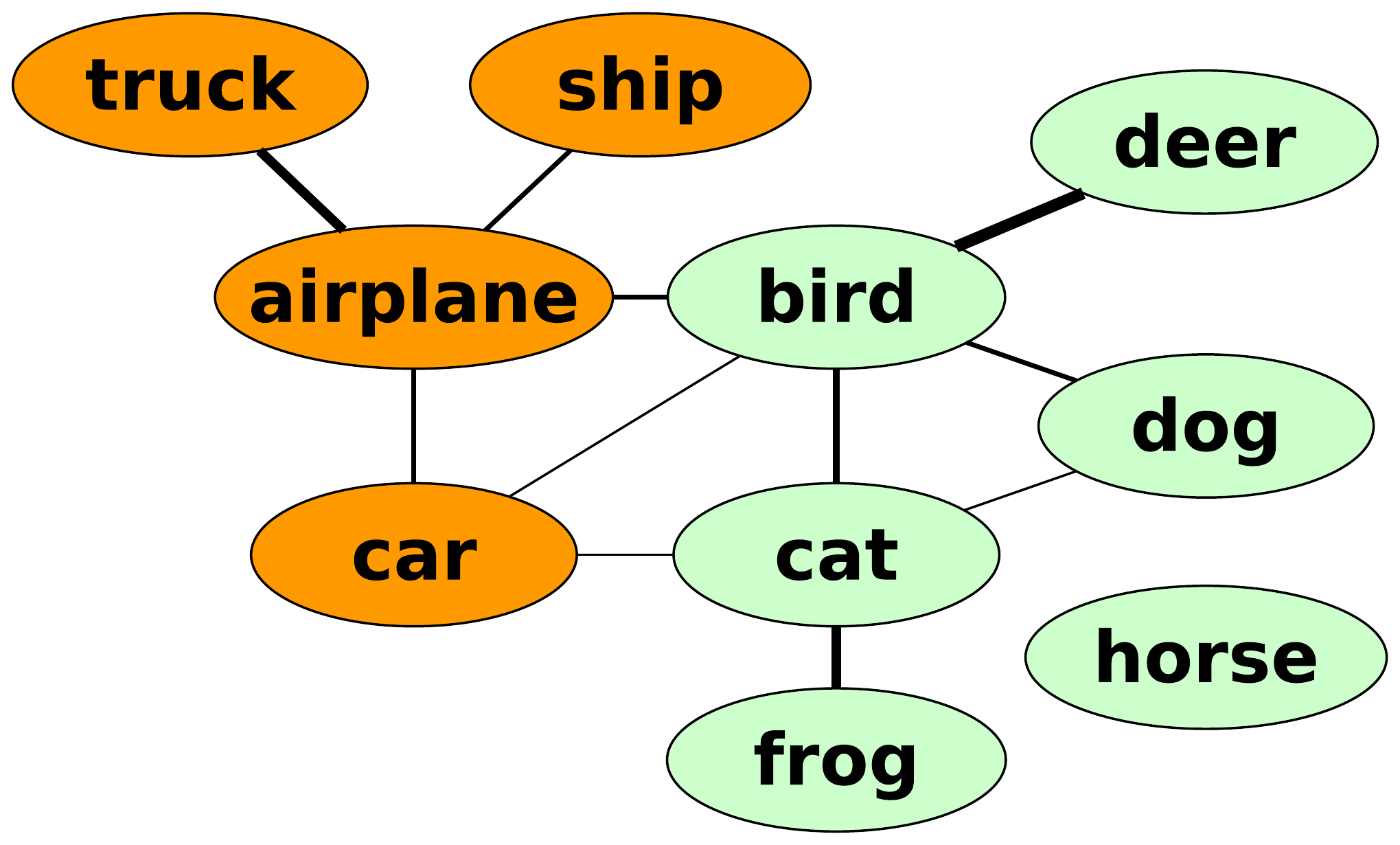}
  	\caption{Relationship graph of CIFAR-10 classes. Edges represent interaction strength between classes.}
    \label{fig:cifar10-graph}
  \end{figure}

  \begin{table}[!t]
  	\caption{Test accuracy${\pm \text{standard deviation}}$ on CIFAR-10 and CIFAR-100 for 
    single task learning (STL), naive MTL (NMTL) and 
    our approach without (HMTL) and with (HMTL-S) the L$1$-norm constraint on matrix $\mc$.}
    \label{tab:cifar10-100}
	\centering
  	\begin{tabular}[b]{lcc}
    \hline
    \abovespace\belowspace
		              & CIFAR-10 & CIFAR-100 \\ 
    \hline
    \abovespace
		STL           & 67.47${\pm \text{2.78}}$ & 18.99${\pm \text{1.12}}$ \\
		NMTL          & 69.41${\pm \text{1.90}}$ & 19.19${\pm \text{0.75}}$ \\
		HMTL		  & 70.85${\pm \text{1.87}}$ & 21.15${\pm \text{0.36}}$ \\ 
		HMTL-S        & 71.62${\pm \text{1.34}}$ & 22.09${\pm \text{0.29}}$ \\
    \hline 
  	\end{tabular}
  \end{table}

Results of five repetitions with different splits are presented in Table \ref{tab:cifar10-100}. Note that HMTL gives a visible improvement in performance, and adding the constraint that $\sum_{j,k} C_{j,k} \leq R$ further improves performance in both datasets. The matrix $C$ can been interpreted as an adjacency matrix of a graph, highlighting the relationships between the classes. Figure \ref{fig:cifar10-graph} depicts the graph for CIFAR-10 extracted from the algorithm HMTL-S. Although this result is strongly influenced by the choice of the data representations, we can note that animals tends to be more related to themselves than to vehicles and vice versa.

\subsection{Phone Classification}
\label{sec:speech}
The aim of the third set of experiments is to assess the efficacy of the
real-time \proc{Forward-HG} algorithm (RTHO). We run experiments on
phone recognition in the multitask framework proposed
in \citep[][and references therein]{badino2016phonetic}. 
Data for all experiments was obtained from the TIMIT phonetic
recognition dataset \citep{garofolo1993darpa}.
The dataset contains 5040 sentences corresponding to around 1.5 million speech acoustic frames. Training, validation and test sets contain respectively 73\%, 23\% and 4\% of the data. 
The primary task is a frame-level phone state classification with 183 classes and it consists in learning a mapping $f_P$ from acoustic speech vectors to hidden Markov model monophone states. Each 25ms speech frame is represented by a 123-dimensional
vector containing 40 Mel frequency scale cepstral coefficients and
energy, augmented with their deltas and delta-deltas. We used a window
of eleven frames centered around the prediction target to create the
1353-dimensional input to $f_P$.
The secondary (or auxiliary) task consists in learning a mapping $f_S$
from acoustic vectors to 300-dimensional real vectors of
context-dependent phonetic embeddings defined in
\citep{badino2016phonetic}. 

  \begin{table}[t]
  \begin{center}
      \caption{Frame level phone-state classification accuracy on 
      standard TIMIT test set and execution time in minutes on one 
      Titan X GPU. For RS, we set a time budget
      of 300 minutes.}
  	\begin{tabular}[b]{lcr}\hline
    \abovespace\belowspace
		& Accuracy $\%$ & Time (min) \\ \hline
        \abovespace
		Vanilla     &  59.81  & 12    \\
        RS & $60.36$ & $300$ \\ 
		RTHO & 61.97  &   164 \\
        RTHO-NT & 61.38   & 289 \\
        \hline 
  	\end{tabular}
  	\label{tab:speech}
  \end{center}
  \end{table}
  
As in previous work, we assume that the
two mappings $f_P$ and $f_S$ share inputs and an intermediate
representation, obtained by four layers of a feed-forward neural
network with 2000 units on each layer. We denote by $W$
the parameter vector of these four shared layers. The network has two different output layers with parameter vectors $W^P$ and $W^S$ each relative to the primary and secondary task. The
network is trained to jointly minimize $
E_{\operatorname{tot}}(W, W^P,W^S) =E_P(W, W^P) + \rho E_S(W, W^S)$, where the
primary error $E_P$ is the average cross-entropy loss on the primary task,
the secondary error $E_S$ is given by mean squared error on the
embedding vectors and $\rho\geq 0$ is a design hyperparameter. Since
we are ultimately interested in learning $f_P$, we formulate the
hyperparameter optimization problem as
$$
\min\big\{ E_{\operatorname{val}}(W_T, W^P_T) ~\text{ subject to }~
\rho,\eta \geq 0, 0 \leq \mu \leq 1\big\},
$$
where $E_{\operatorname{val}}$ is the cross entropy loss
computed on a validation set after $T$ iterations of stochastic GDM,
and $\eta$ and $\mu$ are defined in \eqref{eq:momentum}.
In all the experiments we fix a mini-batch size of 500. 
We compare the following methods:
\begin{enumerate}
\item Vanilla: the secondary target is ignored ($\rho = 0)$; $\eta$ and $\mu$ are set to 0.075 and 0.5 respectively as in \citep{badino2016phonetic}.
\item RS: random search with $\rho\sim\mathcal{U}(0,4)$, $\eta\sim\mathcal{E}(0.1)$ (exponential distribution with scale parameter 0.1) and $\mu\sim\mathcal{U}(0,1)$ \cite{bergstra2012random}.
\item RTHO: real-time hyperparameter optimization with initial learning rate and momentum factor as in Vanilla and initial $\rho$ set to 1.6 (best value obtained by grid-search in \citet{badino2016phonetic}).
\item RTHO-NT: RTHO with ``null teacher,'' i.e. when the initial values of $\rho$, $\eta$ and $\mu$ are set to 0. We regard this experiment as particularly interesting: this initial setting, while clearly not optimal, does not require any background knowledge on the task at hand.
\end{enumerate} 
We also tried to run \proc{Forward-HG} for a fixed number of epochs, not in real-time mode. Results are not reported since the method could not make any appreciable progress after running 24 hours on a Titan X GPU.

\begin{figure}[h]
  \includegraphics[width=90mm, trim={15mm 0cm 0cm 0cm},clip]{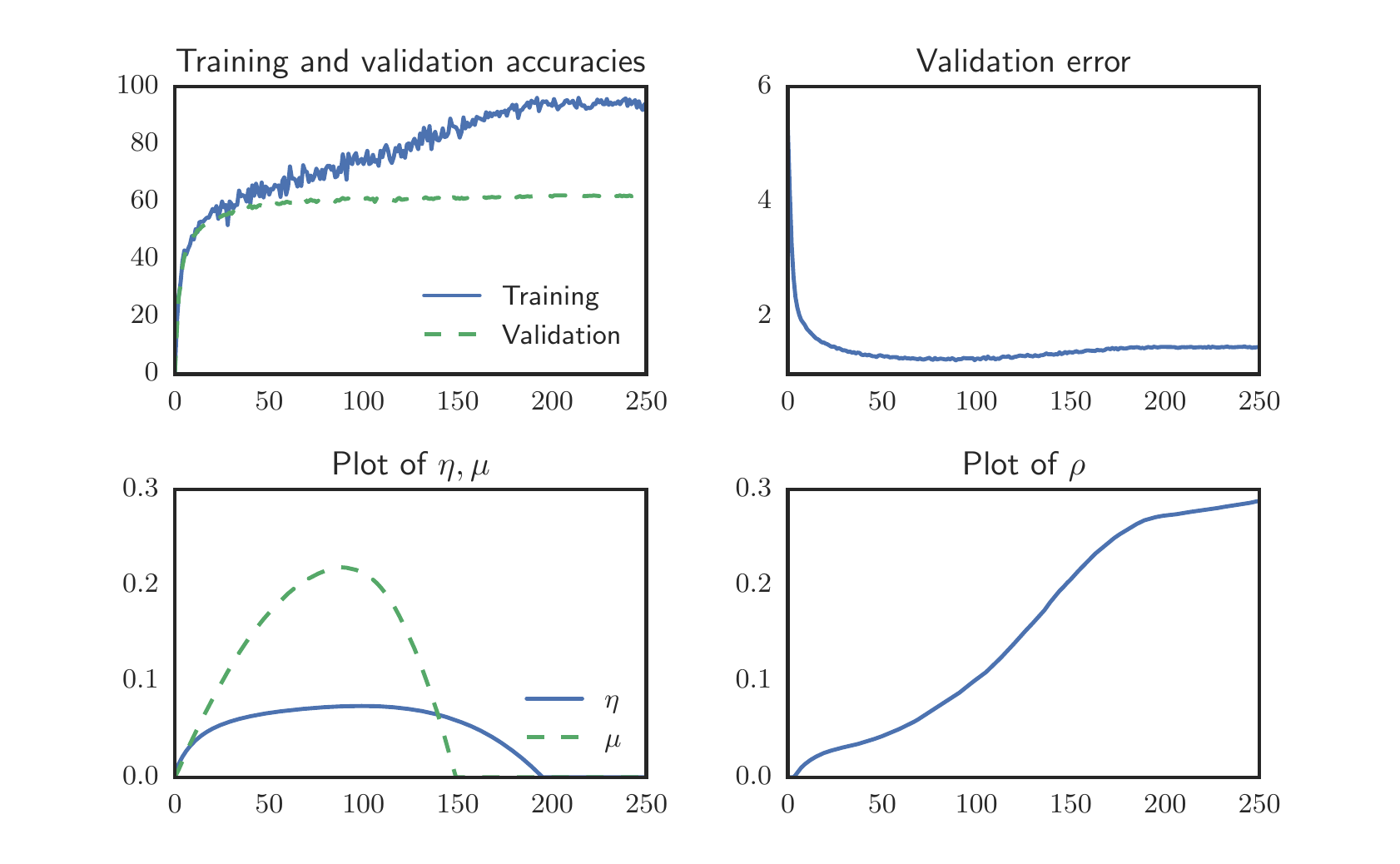}
\caption{Learning curves and hyperparameter evolutions for RTHO-NT: the horizontal axis runs with the hyper-batches. Top-left:
  frame level accuracy on mini-batches (Training) and on a randomly selected
  subset of the validation set (Validation). 
  Top-right: validation error $E_{\operatorname{val}}$ on the same
  subset of the validation set.
  Bottom-left: evolution of optimizer hyperparameters $\eta$ and $\mu$. 
  Bottom-right: evolution of design hyperparameter $\rho$.}
  \label{fig:speech} 
\end{figure}

Test accuracies and execution times are reported in Table \ref{tab:speech}.
Figure \ref{fig:speech} shows learning curves and hyperparameter evolutions for RTHO-NT. 
In Experiments 1 and 2 we employ a standard early stopping procedure on the validation accuracy, while in Experiments 3 and 4 a natural stopping time is given by the decay to 0 of the learning rate (see Figure \ref{fig:speech} left-bottom plot).
In Experiments 3 and 4 we used a 
hyper-batch size of $\Delta=200$ 
(see Eq.~\eqref{eq:partial-hypergradient}) and a hyper-learning rate of 0.005.

The best results in Table~\ref{tab:speech} are very similar
to those obtained in state-of-the-art recognizers
using multitask
learning~\cite{badino2016phonetic,badinopersonal}.
In spite of the small number of hyperparameters,
random search yields results only slightly better than the vanilla
network (the result reported in Table~\ref{tab:speech} are an average
over 5 trials, with a minimum and maximum accuracy of 59.93 and 60.86,
respectively). Within the same time budget of 300 minutes, RTHO-NT is able
to find hyperparameters yielding a substantial improvement over
the vanilla version, thus effectively exploiting the auxiliary task.
Note that the model trained has more that $15\times 10^6$ parameters
for a corresponding state of more than $30\times 10^6$ variables.
To the best of our knowledge, reverse-mode 
\citep{maclaurin2015gradient} or approximate 
\citep{pedregosa_hyperparameter_2016} methods
have not been applied to models of this size.

\section{Discussion}
\label{sec:discussion}
We studied two alternative strategies for computing
the hypergradients of any iterative differentiable learning dynamics.
Previous work has mainly focused on the reverse-mode computation,
attempting to deal with its space complexity, that becomes prohibitive
for very large models such as deep networks.

Our first contribution is the definition and the application 
of forward-mode computation to HO. 
Our analysis suggests that for 
large models the forward-mode computation may be a preferable alternative to reverse-mode
if the number of hyperparameters is small. Additionally, forward-mode
is amenable to real-time hyperparameter updates, which we showed to be an
effective strategy for large datasets (see Section~\ref{sec:speech}). 
We showed experimentally that even starting from a far-from-optimal 
value of the hyperparameters (the null teacher), our RTHO algorithm finds good 
values at a reasonable cost, whereas other
gradient-based algorithms could not be applied in this context.

Our second contribution is the Lagrangian derivation 
of the reverse-mode computation. It provides a general framework
to tackle hyperparameter optimization problems involving a wide class 
of response functions, including those that take into account the whole
parameter optimization dynamics. We have also presented in
Sections~\ref{sec:hypercleaning} and ~\ref{sec:vision}
two non-standard learning problems where we specifically take advantage
of a constrained formulation of the HO problem. 



We close by highlighting some potential extensions of our framework and direction of future research.  First, the relatively low cost of our RTHO algorithm could suggest to make it a standard tool for the optimization of real-valued critical hyperparameters (such as learning rates, regularization factors and error function design coefficient), in context where no previous or expert knowledge is available (e.g. novel domains). Yet, RTHO must be thoroughly validated on diverse datasets and with different models and settings to empirically asses its robustness and its ability to find good hyperparameter values.
Second, in order to perform gradient-based hyperparameter optimization,
it is necessary to set a descent procedure over the hyperparameters.
In our experiments we have always used Adam with a manually
adjusted value for the hyper-learning rate.
Devising procedures which are adaptive in these hyper-hyperparameters is an important direction of future research. 
Third, extensions of gradient-based HO techniques to integer or nominal hyperparameters (such as the depth and the width of a neural network) 
require additional design efforts and may not arise naturally in our framework.
Future research should instead focus on the integration of gradient-based algorithm with Bayesian optimization and/or with emerging reinforcement learning hyperparameter optimization approaches \citep{zoph2016neural}.
A final important problem is to study the converge properties of RTHO. Results in  \citet{pedregosa_hyperparameter_2016} may prove useful in this direction.

\bibliography{Hyperparameters,massi,michele}
\bibliographystyle{icml2017}

\appendix

\section{Empirical Validation Of Complexity Analysis}

\begin{figure}[h]
  \includegraphics[width=\columnwidth]{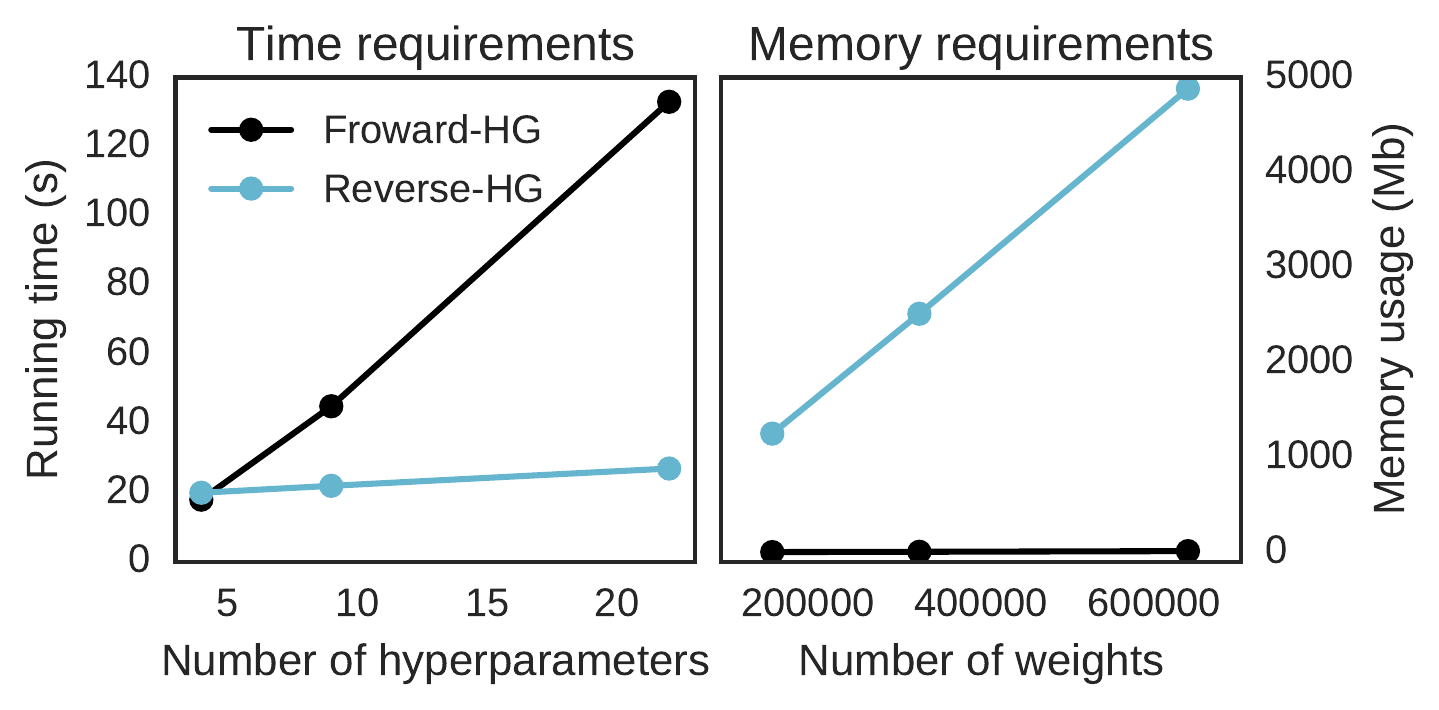}
\caption{Time (left) and space (right) requirements for the computation of the
hypergradient with Forward-HG and Reverse-HG algorithms.}
  \label{fig:time_and_space} 
\end{figure}

To complement the complexity analysis in Section \ref{sec:complexity}, we study empirically the running time per hyper-iteration and
space requirements of Reverse-HG and Forward-HG algorithms. We trained three layers feed-forward neural networks on MNIST dataset with SGDM, for $T = 1000$ iterations. In a first set of experiments (Figure \ref{fig:time_and_space}, left) we fixed the number of weights at 199210 and optimized the learning rate, momentum factor and a varying number of example weights in the training error, similar to the experiment in Section \ref{sec:hypercleaning}. As expected, the running time of Reverse-HG is essentially constant, while that of Forward-HG increases linearly. On the other hand, when fixing the number of hyperparameters (learning rate and momentum factor), the space complexity of Reverse-HG grows linearly with respect to the number of parameters (Figure \ref{fig:time_and_space}, right), while that of Forward-HG remains constant.

\section{Experiments}

\subsection*{Learning Task Interactions}

In Table \ref{tab:mtl-com}, we report comparative results obtained with two
state-of-the-art multitask learning methods (\citet{dinuzzo2011learning} and
\citet{jawanpuria2015efficient}) on the CIFAR-10 dataset.

  \begin{table}[h]
  	\caption{Test accuracy${\pm \text{standard deviation}}$ on CIFAR-10. Hyperparameters of MTL algorithms were validated by grid-search with the same experimental setting of Section \ref{sec:vision}. \citet{jawanpuria2015efficient}'s algorithm contains  a $p$-norm regularizer for the task interaction matrix $C$, for $p\in(1, 2]$. The value of $p$ used in the experiment is specified in the third column.}
    \label{tab:mtl-com}
	\centering
  	\begin{tabular}[b]{lcc}
    \hline
    \abovespace\belowspace
		              & CIFAR-10 & $p$ \\ 
    \hline
    \abovespace
		\citet{dinuzzo2011learning} & 69.96${\pm \text{1.85}}$ & \\
		\citet{jawanpuria2015efficient}          & 70.30${\pm \text{1.05}}$ & $2$ \\
		\citet{jawanpuria2015efficient} & 70.96${\pm \text{1.04}}$  & $4/3$ \\
		HMTL-S        & 71.62${\pm \text{1.34}}$ &  \\
    \hline 
  	\end{tabular}
  \end{table}

Both methods improve over STL and NMTL but perform slightly worse than HMTL-S. The task interaction matrix is treated as a model parameter by these algorithms, which may lead to overfitting for such a small training set, further highlighting the advantages of considering $C$ as an hyperparameter. Computation times are comparable in the order of 2-3 hours. 

Other approaches (e.g. \cite{kang2011learning} tackle the same problem in a similar framework, but a complete analysis of MTL is beyond the scope of this paper.

\subsection*{Phone Classification}

In this section, we present additional experimental results on the phone classification task discussed in Section \ref{sec:speech}. We considered a sequential model-based optimization with Gaussian processes, using the Python package BayesianOptimization found at \url{https://github.com/fmfn/BayesianOptimization/}. We set the following definition intervals for the hyperparameters: $\eta\in[10^{-5}, 1]$, $\mu\in[0, 0.999]$ and $\rho\in[0, 4]$; we used expected improvement as acquisition function and initialized the method with 5 randomly chosen points. In Table \ref{tab:phone-expanded} we report the validation accuracies at different times for this experiment (SMBO) as well as for those presented in Section \ref{sec:speech}.

  \begin{table}[h]
  	\caption{Validation accuracy at different times and final test accuracy for phone classification experiments for various hyperparameter optimization methods.}
    \label{tab:phone-expanded}
	\centering
  	\begin{tabular}[b]{lcccc}
    \hline
    \abovespace\belowspace
		              & 50 min & 100 min & 300 min & Final TA \\ 
    \hline
    \abovespace
		RS & 60.64&60.86&61.23&60.36 \\
SMBO& 60.83 & 60.83 & 61.39&60.91 \\
RHTO-NT& 56.51 & 60.91 & 62.11 & 61.38 \\
RHTO& 59.45 & 61.21 & 62.88 &61.97 \\
    \hline 
  	\end{tabular}
  \end{table}
  
\section{On learning rate initialization in RTHO-NT}
In the RTHO-NT setting presented in Section \ref{sec:speech}, the hyperparameters are initially set to zero. While $\lambda=0$ is, in general, a far from optimal point in the hyperparameter space, it has the advantage of not requiring any previous knowledge of the task at hand.

Here we provide some insights on this particular initialization strategy. For simplicity, we consider the case that the learning dynamics is stochastic gradient descent, that is, $\Phi_t(w_{t-1}, \eta)= w_{t-1} - \eta J_t(w_{t-1})$. In the following, we observe that the first update on the learning rate performed by RTHO-NT is proportional to the scalar product between the gradient of the validation error at $w_0$ and the average of stochastic gradients of the training error over the mini-batches composing the first hyper-batch. Indeed, the partial derivatives of the learning dynamics (see Equation \eqref{eq:AtBt}) are given by
 $$
  A_t = I - \eta H_t(w_0) = I; \quad B_t = - \transpose{[\nabla J_t(w_0)]},
 $$
 where the gradient is regarded as a row vector. Consequently,
 $$
 Z_t = A_t Z_{t-1} + B_t = - \sum_{s=1}^t [\nabla J_s(w_0)]^T
 $$
 and the first hypergradient w.r.t. $\eta$ at time step $t=\Delta$, where $\Delta$ is the hyper-batch size, is 
 $$
 \nabla f_{\Delta}(\eta) = - \nabla E(w_0)  \sum_{s=1}^{\Delta} \transpose{[\nabla J_s(w_0)]}.
 $$
Thus, the smaller the angle between the validation error gradient and the (unnormalized) stochastic training error gradient, the bigger the update. In particular, if the angle is negative, the learning rate would become negative, suggesting that $w_0$ may be a bad parameter initialization point.

\end{document}